\documentclass[acmtog]{acmart}
\acmSubmissionID{258}

\usepackage{booktabs} 
\usepackage[ruled]{algorithm2e} 

\SetAlFnt{\small}
\SetAlCapFnt{\small}
\SetAlCapNameFnt{\small}
\SetAlCapHSkip{0pt}

\usepackage{amsmath,amsthm}

\usepackage{graphicx}
\usepackage{textcomp}
\usepackage{wrapfig}
\usepackage{subfig}
\usepackage{color}
\usepackage{xspace}
\usepackage{subfig}
\usepackage{enumitem}
\usepackage{booktabs}
\usepackage{hyperref}
\usepackage{multirow}
\usepackage{mathrsfs}

\definecolor{blue}{rgb}{0,0,1}
\definecolor{red}{rgb}{1,0,0}
\definecolor{green}{rgb}{0,.5,0}
\definecolor{darkgreen}{rgb}{0,.4,0}
\definecolor{orange}{rgb}{0.75, 0.4, 0}
\definecolor{teal}{rgb}{0.0, 0.4, 0.4}
\definecolor{purple}{rgb}{0.65,0,0.65}
\definecolor{black}{rgb}{0,0,0}

\iftrue
\newcommand{\rz}[1]{{\color{black}\textbf{}#1}\normalfont}
\newcommand{\rh}[1]{{\color{black}\textbf{}#1}\normalfont}
\newcommand{\cl}[1]{{\color{black}\textbf{}#1}\normalfont}

\fi

\usepackage[ruled]{algorithm2e} 

\SetAlFnt{\small}
\SetAlCapFnt{\small}
\SetAlCapNameFnt{\small}
\SetAlCapHSkip{0pt}

\acmJournal{TOG}

\copyrightyear{2025}
\acmYear{2025}
\setcopyright{acmlicensed}\acmConference[SIGGRAPH Conference Papers '25]{Special Interest Group on Computer Graphics and Interactive Techniques Conference Conference Papers }{August 10--14, 2025}{Vancouver, BC, Canada}
\acmBooktitle{Special Interest Group on Computer Graphics and Interactive Techniques Conference Conference Papers (SIGGRAPH Conference Papers '25), August 10--14, 2025, Vancouver, BC, Canada}

\acmDOI{10.1145/3721238.3730610}
\acmISBN{979-8-4007-1540-2/2025/08}

\begin{document}

\title{MASH: Masked Anchored SpHerical Distances for 3D Shape Representation and Generation}

\author{Changhao Li}
\orcid{0000-0003-0850-8987}
\affiliation{%
 \institution{University of Science and Technology of China}
 \city{Hefei}
 \country{China}
}
\email{lch0510@mail.ustc.edu.cn}

\author{Yu Xin}
\orcid{0009-0006-7433-1998}
\affiliation{%
 \institution{University of Science and Technology of China}
 \city{Hefei}
 \country{China}
}
\email{xy0731@mail.ustc.edu.cn}

\author{Xiaowei Zhou}
\orcid{0000-0003-1926-5597}
\affiliation{%
 \institution{State Key Laboratory of CAD \& CG, Zhejiang University}
 \city{Hangzhou}
 \country{China}
}
\email{xwzhou@zju.edu.cn}

\author{Ariel Shamir}
\orcid{0000-0001-7082-7845}
\affiliation{%
 \institution{Reichman University}
 \city{Herzliya}
 \country{Israel}
}
\email{arik@runi.ac.il}

\author{Hao Zhang}
\orcid{0000-0003-1991-119X}
\affiliation{%
 \institution{Simon Fraser University}
 \city{Vancouver}
 \country{Canada}
}
\email{haoz@cs.sfu.ca}

\author{Ligang Liu}
\orcid{0000-0003-4352-1431}
\affiliation{%
 \institution{University of Science and Technology of China}
 \city{Hefei}
 \country{China}
}
\email{lgliu@ustc.edu.cn}

\author{Ruizhen Hu}
\authornote{Corresponding author: Ruizhen Hu (ruizhen.hu@gmail.com)}
\orcid{0000-0002-6798-0336}
\affiliation{%
 \institution{Shenzhen University}
 \city{Shenzhen}
 \country{China}
}
\email{ruizhen.hu@gmail.com}

\begin{abstract}

We introduce {\em Masked Anchored SpHerical Distances\/} ({\em MASH\/}), a novel {\em multi-view\/} and {\em parametrized\/} representation of 3D shapes. 
Inspired by multi-view geometry and motivated by the importance of perceptual shape understanding for learning 3D shapes, 
MASH represents a 3D shape as a collection of {\em observable local surface patches\/}, each defined by a spherical distance function emanating from an anchor point.
We further leverage the compactness of spherical harmonics to encode the MASH functions, combined with a {\em generalized view cone\/} with a parameterized base that 
{\em masks\/} the spatial extent of the spherical function to attain locality.
We develop a differentiable optimization algorithm capable of converting any point cloud into a MASH representation accurately approximating ground-truth surfaces
with arbitrary geometry and topology.
Extensive experiments demonstrate that MASH is versatile for multiple applications including surface reconstruction, shape generation, completion, and blending,
achieving superior performance thanks to its unique representation encompassing both implicit and explicit features.
More information and resources can be found at: \textcolor{blue}{\underline{https://chli.top/MASH}}.

\end{abstract}


%
%


\begin{teaserfigure}
\includegraphics[width=\textwidth]{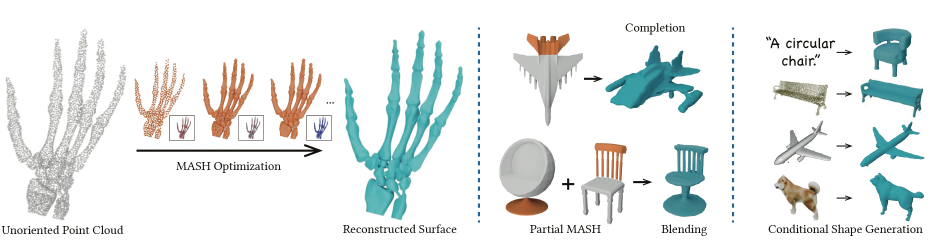}
\caption{MASH represents a 3D shape by fitting a set of  {\em Masked Anchored SpHerical distance functions\/} as observed from the perspective of a fixed number of anchor points in 3D space. Left shows an iterative optimization of MASH parameters, from an \cl{unoriented} point cloud, leading to closer and closer approximations to the ground-truth shape surface. Middle and right show the versatility of MASH in enabling a variety of downstream applications including shape completion, blending, and conditional 3D generation from multi-modal inputs including text prompts, point clouds, and single-view images. On the right: top three results were obtained by training the generator on ShapeNet, while the bottom (the dog) result was obtained on training with Objaverse.}
\label{teaser}
\end{teaserfigure}

\sloppy
\maketitle
\section{Introduction}
\label{sec:intro}

\rz{
Despite a plethora of well-known representations for 3D shapes in vision and graphics, the pursuit after the {\em right representations\/} for targeted tasks remains. In the context of representation learning and generative modeling, the recent success of neural fields~\cite{NF_survey} is well recognized. A key building block of neural fields is} the Signed Distance Function (SDF), which stores a {\em single\/} scalar defining the closest distance to a surface. Although simple and fundamental, SDFs are neither the most informative nor the most efficient since many samples have to be taken to closely define a surface.


From the perspective of multi-view geometry and treating each point in space as a viewpoint, a large \rz{and rich} amount of surface information can be {\em observed\/}. As we scale up per-point information, we can reduce the number of points (views) needed to capture the underlying surface, striving for a better tradeoff between representational capacity and compactness. Taking such a {\em perceptual\/} approach to represent 3D shapes accentuates shape {\em understanding\/}. This is critical to any learning task that relies on such an understanding to compensate for the ill-posed nature of the task due to sparse inputs. Primary examples of such tasks include single-view 3D reconstruction, shape completion, and text-to-3D generation.

In this work, we introduce a novel {\em multi-view\/} and {\em parametric\/} representation of 3D shapes called \emph{Masked Anchored SpHerical (MASH) distances}. Inspired by multi-view geometry, MASH represents a 3D shape as a collection of {\em observable local surface patches\/}, each defined by a spherical distance function emanating from an anchor point of observation.
The creation of such a representation is not straightforward since one has to determine how to parametrize the spherical distance functions and the precise demarcation of their view ranges. Achieving a balance is paramount, \rz{as} we aim to minimize the overlap between local regions associated with different anchors while ensuring that the local connectivity between adjacent patches remains intact.
To this end, MASH leverages the mathematical elegance of {\em spherical harmonics\/}, combined with a {\em generalized view cone\/} \rz{to form a} parameterized base that {\em masks\/} the spatial extent of the spherical function; see Figure~\ref{fig:mash}. 




Our new 3D shape representation presents several distinct features. First, each spherical function in MASH is significantly more informative than a single closest distance as in an SDF, while the need for much fewer spherical functions (about 400) and the use of spherical harmonics enable a compact representation that maintains local surface smoothness and continuity. Second, the masking technique ensures that each patch can be accurately approximated by spherical harmonics and provides a balance among patches to avoid computational redundancy. 
To further ensure efficient optimization and adaptation to geometric details, we introduce a differentiable optimization scheme for MASH based on its parameterized representation, which consists of several novel key components including differentiable point sampling, anchored inversion transformation, and a two-step optimization strategy with tailored loss functions. 

As a compact, parametrized, and patch-based representation, MASH offers several unique advantages in downstream applications and
its versatility is at display in Figure~\ref{teaser}.
To start, MASH is inherently equipped to handle complex topologies and can be iteratively refined to closely approximate 3D shapes from even un-oriented point clouds.
Additionally, being parameterized, MASH provides a natural embedding for 3D shapes and thus can be used for shape generation, with the generative model producing MASH representations directly instead of some intermediate, typically non-interpretable, implicit features. 
The compactness introduced by the MASH parametrization as well as the fact that different shapes can share similar {\em local\/} surface patches makes our MASH-based generative models easier to train and converge faster than alternative representations.
Moreover, being patch-based hence with local support, MASH further allows explicit editing of the generated shapes, which is difficult to achieve with purely implicit representations. 
More specifically, any subset of the anchors and thus its associated surface patches can be frozen during shape generation to facilitate applications such as shape completion or blending.

Our main contributions can be summarized as follows:
\begin{itemize}
\item MASH, a novel multi-view and parametrized representation for 3D shapes that accurately captures the surfaces of objects with arbitrary geometry and topology. \rz{MASH is discriminative yet compact, leading to more efficient learning and inference for generative tasks compared to existing neural implicit shape representations~\cite{NF_survey}.}
\item A differentiable MASH optimization capable of 
converting any input point cloud into a precise MASH representation.
\item \rz{MASH enables multiple applications including shape generation, completion, blending, and reconstruction. For the latter, the locality of the explicit patches and the high representability and smoothness of each parameterized patch enable MASH to better capture local details.}
%

\end{itemize}

Extensive experiments validate the effectiveness of MASH in reconstructive and generative tasks over traditional and \rz{prior implicit representations, achieving superior performance thanks to the inclusion of both implicit and explicit features in MASH.}

\section{Related Work}

\paragraph{3D Shape Representations}
Point clouds, meshes, and voxels have been the most adopted explicit 3D representations. 
While point clouds can be high-resolution, their unordered nature makes them difficult to process. Triangle meshes deliver more detailed surface geometr and are favored for tasks that require more accurate surface representations\cl{~\cite{lim2018simple, gong2019spiralnet, Hanocka_2019, sun2024review}}, but their complex topologies make them harder to encode and generate. Voxel grids, as a natural extension from 2D pixels, are easier to process thanks to their rasterized nature and are thus widely used for shape reconstruction and generation~\cite{wu20153d, choy20163d, girdhar2016learning, wu2016learning, brock2016generative, dai2017shape}. However, their memory consumption grows exponentially as the resolution increases. 

Implicit\cl{~\cite{calakli2012ssdc, hoppe1992surface, kazh2006poisson, 
pgr, sun2024recent}}, \rz{especially} neural implicit, representations have become popular. \rz{The latter} encode shapes using continuous functions such as SDFs \cite{park2019deepsdf, calakli2012ssdc} or occupancy fields \cite{mescheder2019occupancy,chen2019learning,conv-onet}, offering \rz{continuity}, flexibility, and the ability to represent complex topologies smoothly. However, their implicit nature \rz{hinders} explicit shape editing. Parametric methods such as spline surfaces \cite{gordon1974b, iglesias2004functional} and spherical harmonics \cite{saupe20013d} use compact representations but \cl{struggle to obtain suitable parametric representations to accurately describe arbitrary complex geometries.}

Our MASH representation combines the advantages of \rz{the above} representations, with the set of anchors \rz{offering the flexibility} of point clouds, the surface patches parametrized with each anchor providing more detailed geometry, and the parametrization itself \rz{reflecting} a natural implicit embedding for 3D shapes.

Among prior works, ARO-Net~\shortcite{aronet} is most related \rz{to MASH since} it also utilizes partial observations at a set of anchors, \rz{with the goal of improving surface reconstruction from sparse point clouds}. However, their observation is oriented to a query point for occupancy prediction and there is no parameterization or optimization involved. 
\rz{In our work,} we obtain an explicit parametric representation by optimizing MASH directly from a point cloud. \rz{Then a surface can be extracted from the MASH representation.}

\paragraph{3D Shape Generation}
Recent advances in 2D image generation, such as DALL$\cdot$E \cite{ramesh2021zero}, Imagen \cite{Imagen2022}, and Stable Diffusion \cite{rombach2022high}, have inspired 3D generation methods that leverage 2D priors.
DreamFusion \cite{poole2022dreamfusion} introduces Score Distillation Sampling (SDS) to optimize 3D shapes via NeRF \cite{mildenhall2021nerf}, and some works focus on extending the concept of SDS to various neural domains \cite{liu2023zero, li2023instant}.
However, these methods suffer from a long optimization time and \cl{geometric inconsistencies in multi-view images may lead to geometric artifacts.}
Another line of work for 3D shape generation is to directly train and generate 3D representations. 
Early methods \cite{choy20163d, mescheder2019occupancy} primarily utilize 3D convolutional networks to encode and decode 3D voxel grids.
Point-E \cite{nichol2022point} innovatively employs a diffusion model based on the pure transformer network structure to directly generate point clouds.
Polygen \cite{nash2020polygen} and MeshGPT \cite{siddiqui2024meshgpt} proposed to generate meshes by serializing the vertices and faces of a mesh, 
generating high-quality results, but their reliance on high-quality datasets limits their generality.
Later, with the advent of Variational Autoencoder (VAE), many works \cite{cheng2023sdfusion, gupta20233dgen, jun2023shap, zhang20233dshape2vecset, zhao2024michelangelo} use VAE to encode 3D shapes and decode them into occupancy or distance fields, and generate shapes in the encoded latent space.
Unlike those methods, we adopt the explicit parametrized MASH representation directly for generation, 
improving efficiency while maintaining accuracy.


\section{Method}


In this section, we first explain the parametric representation of MASH in Section~\ref{subsec:mash}, and then show how to optimize the MASH representation for a given shape in Section~\ref{subsec:opt}. 

\subsection{MASH Representation}
\label{subsec:mash}

\begin{figure}[!t]
\includegraphics[width=0.48\textwidth]{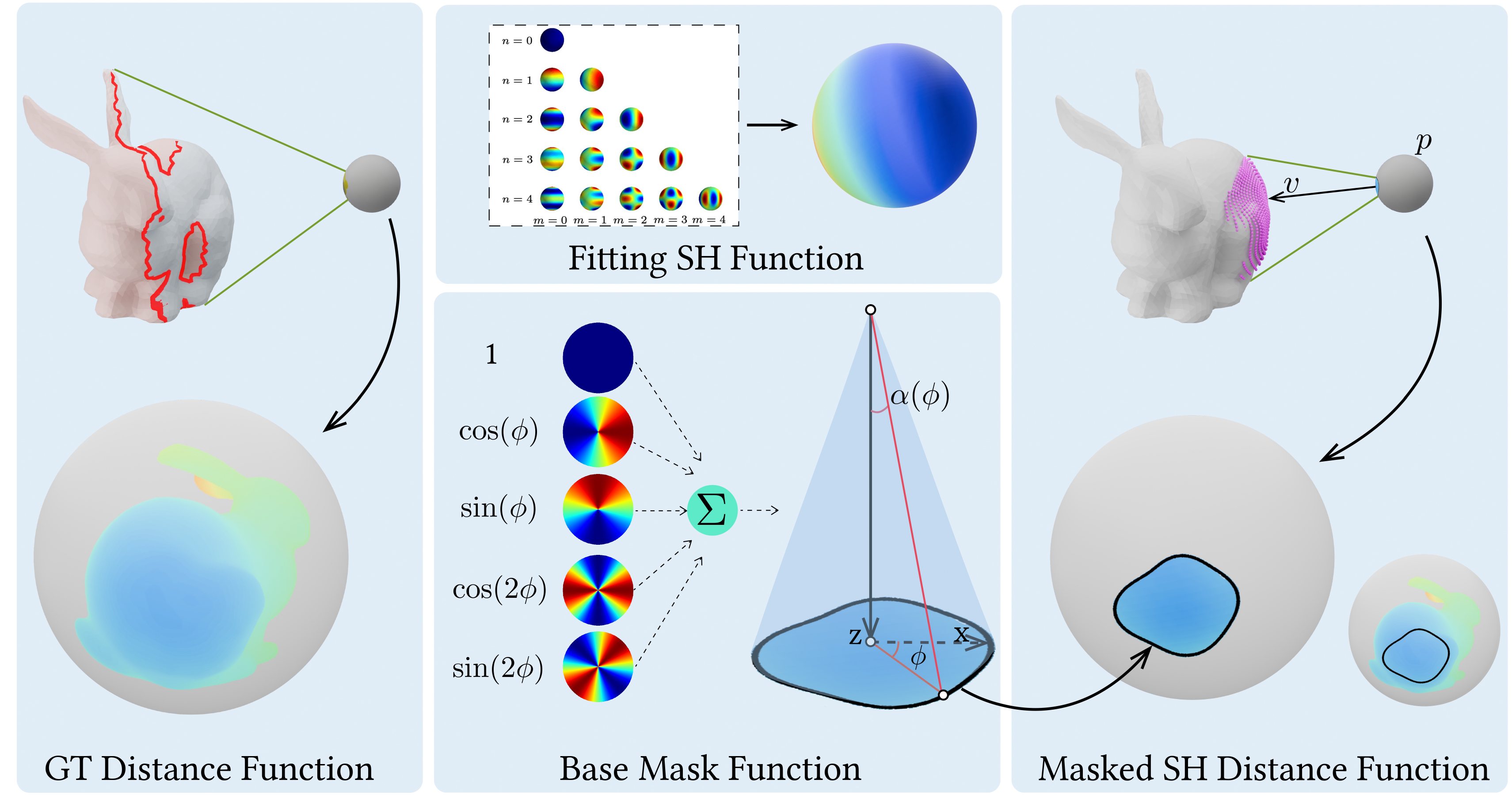}
\caption{Parametric representation of MASH for a single anchor.
}
\label{fig:mash}
\end{figure}

Given a 3D shape $S$ and a point $p$ in space, we can define the visible region on the surface of $S$ from the perspective of $p$ by casting rays from $p$ in all directions, forming a spherical distance function centered at the anchor point $p$, as shown in Figure~\ref{fig:mash} (left).
Such visible regions have been studied for shape partition~\cite{SDF}, reconstruction~\cite{SSZCO-10}, and relationship optimization between two shapes~\cite{zhao2016relationship}, all suggesting that visible regions can accurately characterize local shape features, while global structures of a 3D shape can be faithfully captured from a {\em set } of anchor points all around the surface of the shape. 

Our goal is to define a parameterized representation of these visible regions such that a 3D shape can be represented by a set of anchored parameters, with a simple and compact structure. To this end,
we use a set of spherical harmonics \rz{(SH)} to approximate the visible region represented by a spherical distance function. However, the spherical distance function is discontinuous when shifting from one surface region to other unconnected regions, as well as when crossing view boundaries. Such discontinuities hinder the use of SH for an accurate approximation of an entire distance function.
The key idea of MASH is to further introduce a {\em mask\/} 
to constrain the approximation region so that even low-order SH can provide a faithful approximation. 
Specifically, the mask is defined by a generalized 3D view cone with a parameterized free-form base. 

Therefore, each anchor of our MASH can be represented by a set of parameters $\mathcal{A} = \{p, v, \mathcal{C}, \mathcal{V}\}$, where $p$ is the location of the anchor, $v$ refers to the view direction, $\mathcal{C}$ and $\mathcal{V}$ are two subsets of parameters used to define the corresponding spherical distances and vision mask.
Figure~\ref{fig:mash} shows our MASH representation for a single anchor point. 
The two subsets of parameters $\mathcal{C}$ and $\mathcal{V}$ are visualized in the middle, and the continuous visible region characterized by our MASH representation is shown on the right.


\paragraph{Parameterized spherical distances.}
We use a combination of SH to represent the spherical distance function as follows:
\begin{equation}
\label{eq:sh}
    d_{\mathcal{C} }(\theta,\phi) = \sum_{l=0}^{l=L}\sum_{m=-l}^{m=l} {C_l^m Y_l^m(\theta,\phi) },
\end{equation}
where $Y_l^m$ is the SH at frequency $l$ and $C_l^m$ are the corresponding combination coefficients. Thus, the SH parameters are defined as 
\begin{equation}
    \mathcal{C} = \{C_l^m \big| |m| \leq l,  l=0,1,\dots,L \}.
\end{equation}


\paragraph{Parameterized vision mask.}
We use a generalized 3D view cone to constrain the vision field and define the mask on the sphere.
Note that the common right circular cone with a fixed angle $\alpha$ can only define a circular region on the sphere, which cannot characterize the free-form surface boundary obtained when viewing from a certain direction as shown in Figure~\ref{fig:mash} (left).
Therefore, we generalize the 3D view cone with an anisotropic vision angle $\alpha(\phi)$, where $\phi \in [0, 2\pi]$ defines a circle around the Z-axis direction and $\alpha(\phi)$ is the maximal view range corresponding to each $\phi$, as shown in Figure~\ref{fig:mash} (middle).

Note that $\alpha(\phi)$ is a periodic function, thus it can also be approximated by trigonometric interpolation as follows:
\cl{
\begin{equation}
\label{eq:mask}
    \alpha(\phi) = \pi\sigma(a_0 + \sum_{k=1}^{K} {a_k \cos{(k\phi)}} + \sum_{k=1}^{K} {b_k \sin{(k\phi)}}),
\end{equation}
}
where
\begin{equation}
    \sigma(x) = \frac{1}{1 + e^{-x}}.
\end{equation}
Thus, the vision mask can be defined by the set of parameters:
\begin{equation}
    \mathcal{V} = \{a_0, a_1, b_1, \dots, a_K, b_K\}.
\end{equation}

\subsection{MASH Optimization}
\label{subsec:opt}


Given a 3D shape $S$, either a mesh or a point cloud, the goal of MASH optimization is to find a set of anchors with associated parameters $\{\mathcal{A}_i\}_{i=1}^{M} = \{p_i, v_i, \mathcal{C}_i,  \mathcal{V}_i\}_{i=1}^{M}$ to attain high approximation quality.

To enable {\em differentiable\/} optimization of $\{\mathcal{A}_i\}$, we first \cl{sample points from the surface patches defined by our MASH representation, then formulate a differential operator to calculate the sample point coordinates from the MASH parameters.} \rz{Chamfer Distances (CDs) between the \rz{sample} points and points from the given shape define the approximation error and guide the optimization.}

Moreover, note that for each anchor, intuitively, $\mathcal{V}$ defines a local patch on a sphere, and $\mathcal{C}$ characterizes the local geometry of such surface patch, as shown in Figure~\ref{fig:mash} 
 (right). As $\mathcal{C}$ is a set of SH parameters, it can be used to represent a spherical surface more precisely than a \cl{planar} surface. To further boost the \cl{representability} of the SH, we also introduce the inverse transformation \cite{katz2015visibility} that can convert planner surfaces of the given shape into spherical surfaces and thus make them easier to fit.


\begin{figure}[!t]
\includegraphics[width=0.45\textwidth]{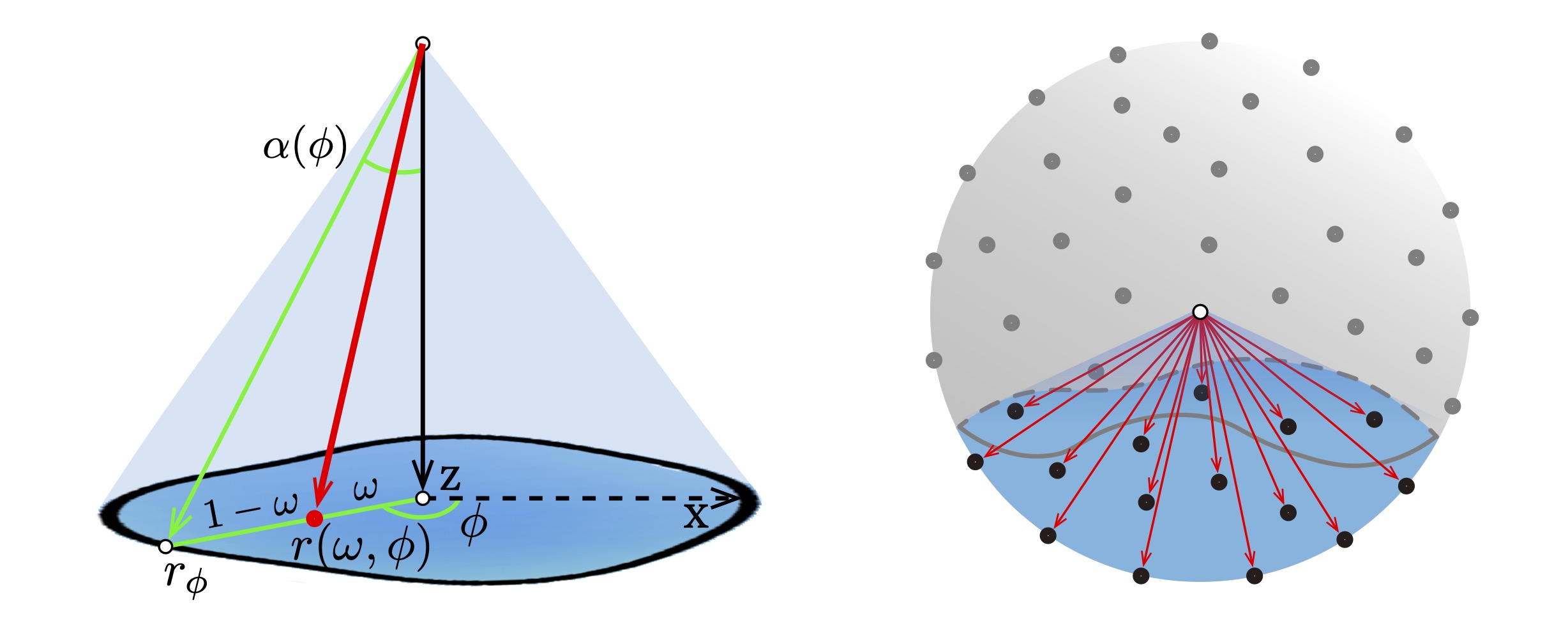}
\caption{Differentiable MASH point sampling. 
Parametrization of any ray sampled inside the view cone (left) and pre-sampled rays filtered by the view cone (right), with each ray corresponding to one sample point.
}
\label{fig:sample}
\end{figure}

\paragraph{Differentiable point sampling.}
To sample points on the \rz{surface patch defined by each anchor is essentially the same as sampling} rays inside the corresponding 3D vision cones, as each ray intersects the surface patch at one point. 
Moreover, each ray also intersects the base mask at one point and can be represented by two parameters $\{ \omega, \phi \}$ inside the base mask as shown in Figure~\ref{fig:sample} (left).

More specifically, $\phi\in[0, 2\pi]$ defines a point on the mask boundary, and $\omega\in[0, 1]$ determines the interior point on the line segment connecting the boundary point and the center, and the ray shooting from the anchor to that interior point be calculated by the spherical linear interpolation \cite{pennec1998computing}:
\begin{equation}
\label{eq:dir-inter}
    r(\omega, \phi) = \text{slerp}(\mathbf{z}, r_{\phi}, \omega),
\end{equation}
where $\mathbf{z}$ is the Z-axis direction and $r_{\phi}$ is the ray shooting from the anchor to the boundary point determined by $\phi$:
\begin{equation}
    r_{\phi} = (\sin(\alpha(\phi)) \cos\phi, \sin(\alpha(\phi)) \sin\phi, \cos(\alpha(\phi))),
\end{equation}
with $\alpha(\phi)$ defined by Equ.~(\ref{eq:mask}).

Now with the sampled ray $r(\omega, \phi)$, the corresponding intersecting point on the surface patch can be obtained by moving the anchor along the ray with the corresponding spherical distance $d(\alpha(\phi),\phi)$ defined by Equ.~(\ref{eq:sh}).
As all the computation so far is conducted in the local coordinate of the anchor, to get the final position of the points on the surface, we still need to apply transformation determined by the position $p$ and view direction $v$ of the anchor. Therefore, the final surface point can be obtained as follows:
\cl{
\begin{equation}
    \hat{p}(\omega, \phi) = p + \mathbf{R}_v \cdot d(\omega \alpha(\phi),\phi) \cdot r(\omega, \phi),
\end{equation}
}
where  $\mathbf{R}_v$ is the rotation matrix determined by $v$ and defined as: 
\begin{equation}
    \mathbf{R}_v = \cos\theta_v\mathbf{I} + (1 - \cos\theta_v) kk^T + \sin\theta_v \mathbf{K},
\end{equation}
with
\begin{equation}
    k = \frac{v}{\|v\|}, \quad \theta_v = \|v\|, \quad
    \mathbf{K} = \begin{bmatrix}
    0 & -k_z & k_y \\
    k_z & 0 & -k_x \\
    -k_y & k_x & 0
    \end{bmatrix}.
\end{equation}
Note that to uniquely determine $\mathbf{R}_v$ from $v$ as above, $v$ does not directly record the exact view direction of the anchor, but instead stores the rotation axis (in direction $k$) and rotation angle (in magnitude $\theta_v$), which are used to rotate the local coordinate of the anchor so that the Z-axis is pointing to the view direction.

To sample a set of rays inside the view cone, one straightforward way is to uniformly sample $\omega \in [0, 1]$ and $\phi \in [0, 2\pi]$. However, this kind of sampling usually leads to non-uniform distribution of corresponding sample points on the surface patch, with local regions staying farther to the anchor get sparse points. To ensure more uniform sampling on the surface patches, we pre-sample uniform points on a unit sphere and then select the subset of points inside our view cone for the shooting rays, as shown in Figure~\ref{fig:sample} (right). 

In more detail, we first uniformly sample $N_{\text{dir}}$ points on the unit sphere using Fibonacci sampling \cite{keinert2015spherical}. Therefore, we have a set of pre-defined ray directions parametrized in the spherical coordinate $\{ \theta^{\text{pre}}_j, \phi^{\text{pre}}_j \}$, with
\begin{equation}
\label{eq:theta}
    \theta^{\text{pre}}_j = \arccos(1 - \frac{2 \cdot j - 1}{N_{\text{dir}}}), j \in [1, N_{\text{dir}}],
\end{equation}
\begin{equation}
\label{eq:phi}
    \phi^{\text{pre}}_j = (1 + \sqrt{5}) \cdot \pi \cdot (j - 0.5), j \in [1, N_{\text{dir}}].
\end{equation}
Then, we can convert those two parameters into the local coordinate of our base mask $\{ \omega^{\text{pre}}_j, \phi^{\text{pre}}_j \}$ with $\omega^{\text{pre}}_j = \theta^{\text{pre}}_j / \alpha(\phi^{\text{pre}}_j)$, and then filter the parametric pre-sampled directions with $\omega^{\text{pre}}_j \in[0,1]$.

\cl{
It is worth noting that after each iteration of our MASH optimization, the updating of the vision mask may lead to a change in the number of filtered pre-sampled rays.
Therefore, at the beginning of each iteration, we obtain the rays within the vision mask using the aforementioned method for subsequent differentiable calculations. Meanwhile, this process itself is non-differentiable.
}

\begin{figure}[!t]
\includegraphics[width=0.46\textwidth]{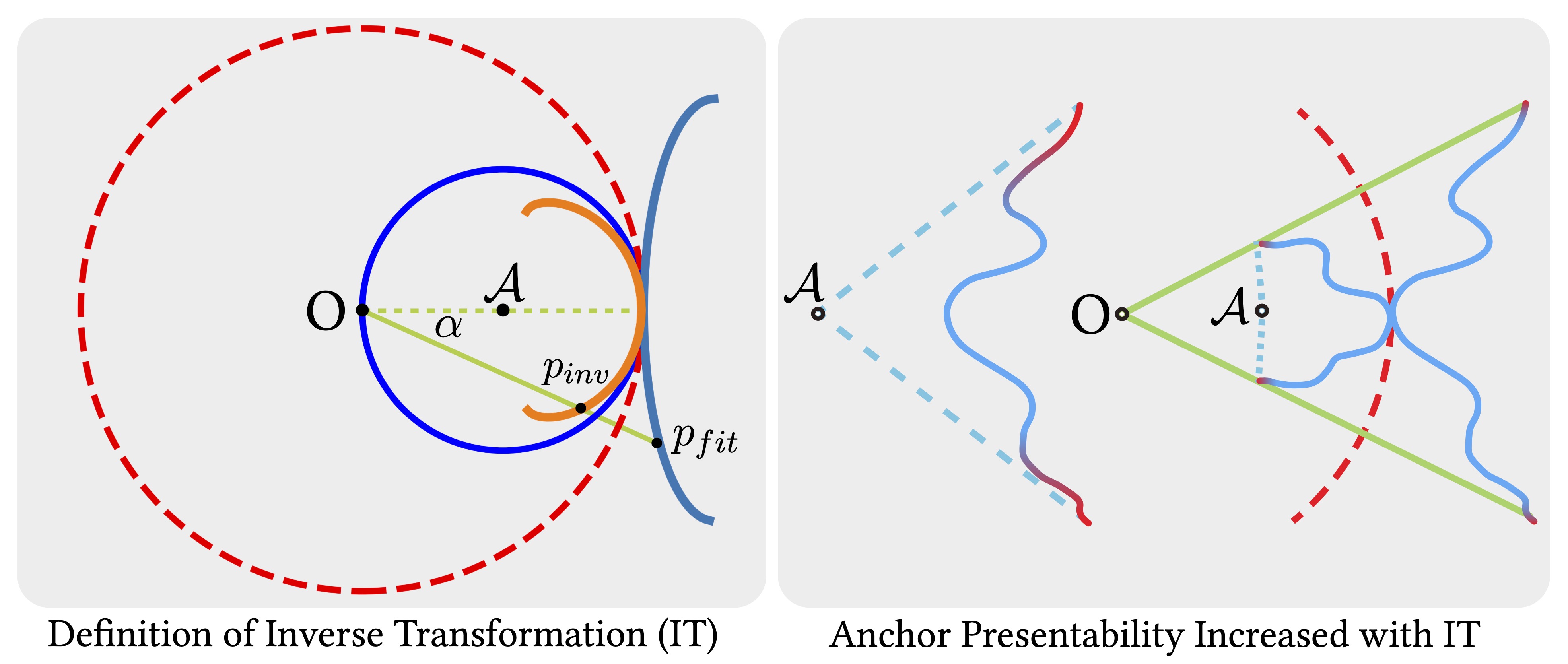}
\caption{Anchored inverse transformation. }
\label{fig:inverse}
\end{figure}

\paragraph{Anchored inverse transformation.}
Our goal is to enlarge the visible region of each anchor when given the fixed degree of the SH and approximation threshold, such that we can use fewer anchors to reach to the same level of approximation. 
Figure~\ref{fig:inverse} \rz{(left) visually defines the inverse transformation} and how the anchor is able to approximate the given surface after applying such a transformation (right). The curve shown on the right represents the surface that needs to be approximated with MASH, and the colormap on top of it indicates the fitting error, with red corresponding to a larger error. After inverse transformation, this surface is bent around the anchor, making it easier for SH to fit and leading to lower fitting error, which results in a larger fitting region to the anchor.


For a more formal definition of the inverse transformation, we first define the center of the inverse transformation $O$, within the local coordinate of each anchor, as:
\begin{equation}
    O = - h \cdot \mathbf{z},
\end{equation}
where $h$ is the distance from the anchor along $\mathbf{z}$ to its parametric surface, which can be approximated by the first SH parameter $C_0^0$ for efficiency. 
Next, we \rz{define} the \rz{relations} between the target fitting point $p_{\text{fit}}$ on the source surface and the target inverse point $p_{\text{inv}}$:
\begin{equation}
    p_{\text{inv}} = O + \frac{R^2}{||p_{\text{fit}} - O||_2^2} \cdot (p_{\text{fit}} - O),
\end{equation}
\begin{equation}
\label{eq:inv}
    p_{\text{fit}} = O + \frac{R^2}{||p_{\text{inv}} - O||_2^2} \cdot (p_{\text{inv}} - O),
\end{equation}
where $R=2C_0^0$ is the radius of the inverse sphere.
We apply the inverse transformation to the parametric points sampled on the patch surfaces using Equ.~(\ref{eq:inv}) before rotating and translating them.

\paragraph{Loss functions.}


Given the points $Q$ sampled on the source shape and the parametric points $P$ sampled on the MASH patches, we now define the loss functions to guide the optimization of our MASH parameters.
First, we utilize the \rz{L1-CD} and split the two main terms as the fitting loss term $L_{f}$ and coverage loss term $L_{c}$ individually:
\begin{equation}
\label{eq:loss-fit}
    L_{f}(P, Q) = \frac{1}{|P|} \mathop{\sum}\limits_{p \in P} d(p, Q),
\end{equation}
\begin{equation}
\label{eq:loss-cover}
    L_{c}(P, Q) = \frac{1}{|Q|} \mathop{\sum}\limits_{q \in Q} d(q, P),
\end{equation}
where the point-to-set distance is defined as $d(p, Q) = \mathop{\min}\limits_{q \in Q}||p - q||_2$.
Note that $L_{f}$ will make the sampled points $P$ to be closer to the input point cloud $Q$, while $L_{c}$ will make the sampled points $P$ cover as more points in $Q$ as possible.
Thus, we can control the moving trend of $P$ by setting different weights for these two loss items.

However, these loss terms only measure the distances between discrete point sets, which encourage anchors to cover the input point cloud, neglecting the continuity between surfaces corresponding to adjacent anchors.
Therefore, we propose a new boundary-continuous loss function to improve the connectivity of anchor mask boundaries, 
which is defined as:
\begin{equation}
\label{eq:loss-bc}
    L_{b}(P) = \frac{1}{M} \mathop{\sum}\limits_{i \in [1, M]} \frac{1}{|P_{i}|} \mathop{\sum}\limits_{p \in P_{i}} d(p, \mathop{\cup}\limits_{j \neq i} P_{j}),
\end{equation}
where $M$ is the number of anchors, $P_{i}$ the sample point on the mask boundary of the $i$-th anchor.
Our full loss function is then:
\begin{equation}
\label{eq:loss}
    L = \omega_{f} L_{f} + \omega_{c} L_{c} + \omega_{b} L_{b}.
\end{equation}
Note that all the equations we have defined so far have analytical expressions.
\cl{After filtering out the rays represented as $\{ \omega^{pre}_j, \phi^{pre}_j \}$ within the view cone of each anchor, } we can calculate the derivatives of all MASH parameters with respect to the final loss value by composing these functions using the Chain Rule for differentiation. Therefore, our whole optimization scheme is differentiable.

\begin{figure}[!t]
\includegraphics[width=0.45\textwidth]{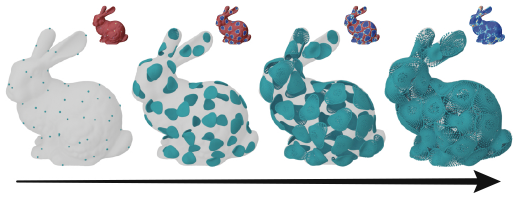}
\caption{
MASH optimization process.
}
\label{fig:fitting}
\end{figure}

\paragraph{Optimization process.}
We adopt a two-step optimization strategy which first ensures a full coverage of the source shape and then refines the local details. The optimization process for the Bunny with $M=50$ anchors is illustrated in Figure~\ref{fig:fitting}, with the fitting error on each step shown on the top right corner.

For initialization, we first uniformly sample $M$ points from the source point cloud $Q$, and then move those points along their locally estimated normal direction with a pre-defined small distance $d_{\text{init}}$ to get the initial locations of all the anchors, with each anchor pointing to the corresponding sampled point. All the remaining parameters are set to 0, except $ C_0^0=d_{\text{init}}/Y_0^0$, to initialize the surface patch corresponding to each anchor as a planar disk close to the source shape after inverse transformation.

Note that the boundary-continuous loss $L_b$ defined in Equ.~(\ref{eq:loss-bc}) is mainly used to improve the boundary continuity between anchor patches, so it is only introduced when the main task of approximating the source shape is completed. Therefore, in the first stage of the optimization, we set $\omega_{b} = 0$ to ensure that anchors can expand along and cover the source shape, with $\omega_{f} = 1$ and $\omega_{c}$ linearly increasing from 0.5 to 1. 
Once 80\% of the source shape is covered, we enter the second stage of the optimization by linearly increasing $\omega_{b}$ from 0 to 1, to ensure that each anchor continues to prioritize fitting the object surface as its main objective and at the same time enhance the connectivity between adjacent anchors.
The optimization continues until the loss $L$ defined in Equ.~(\ref{eq:loss}) converges.



\begin{figure}[!t]
\includegraphics[width=0.49\textwidth]{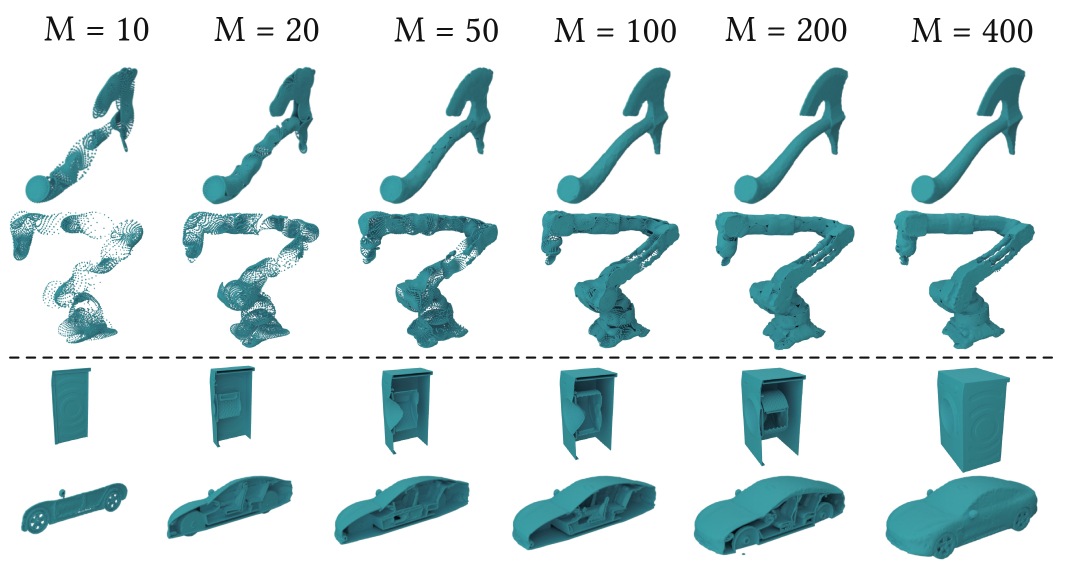}
\caption{MASH optimization results with different anchor numbers (top) and for shapes with inner or thin structures (bottom).
}
\label{fig:diff_M}
\end{figure}

\section{Experiments and Results}


In this section, we first show the expressive power of the proposed MASH representation in Section~\ref{sec:rep}, and then demonstrate how MASH can be used in two different applications, including surface reconstruction in Section~\ref{sec:recon} and shape generation in Section~\ref{sec:gen}.

\subsection{Shape Approximation}
\label{sec:rep}

The key hyperparameters in our MASH representation are anchor numbers $M$, mask degrees $K$, and SH degrees $L$, and the total number of MASH parameters $N$ can be calculated as 
    $N = M (2 K + 1 + (L + 1)^{2} + 6)$. 
The representation ability of MASH will grow with the increase of $M$, $K$, and $L$. However, a larger number of parameters will lead to a reduction in computational efficiency. 
 In all our experiments, we set $M=400$, $K=3$, and $L=2$.
 Under this setting, it takes an average of about 39 seconds \cl{on a single RTX 4090} to fit a single object and the calculation of the loss terms takes approximately 80\% of the total time.
 More quantitative comparisons of the results with different hyperparameters and the effect of applying inverse transformation can be found in the supplementary material. 

 Figure~\ref{fig:diff_M} (top) shows some MASH optimization results with different anchor numbers, with the surface patch represented by each anchor visualized using a set of sampled points. We can see that even with only 10 anchors, MASH can faithfully capture shape with a set of bent patches. With the increase of anchor number $M$, the approximate error becomes smaller, and the sharp features are better preserved. \cl{For example, we can see that the flat bottom of the axe is well approximated using one anchor when $M = 10$, and for the axe blade, to preserve the shape feature, our optimization tends to place anchors on both sides and use the mask boundary to fit the feature.} Note that there is \cl{no assumption} on the topology of the given shape, and MASH can be used to approximate shapes with inner or thin structures as shown in Figure~\ref{fig:diff_M} (bottom). 



\subsection{Surface Reconstruction} 
\label{sec:recon}

\rz{
To verify that our patch-based MASH representation can effectively capture local details, due to its locality and the high representability and smoothness of each parameterized patch, we implemented an efficient surface extraction method, coined M+M, to convert MASH into a mesh for fair and extensive comparisons.}

\paragraph{MASH surface extraction}
Given a point cloud, we can optimize its MASH representation as described in Section~\ref{subsec:opt}. 
To further obtain a watertight surface, 
we first estimate the normals on points $P$ sampled on our surface patches and then extract the iso-surface with marching cubes on octrees with a maximum depth 10, similar to \cl{the proposed iso-surfacing method in PGR \cite{pgr}}. 

Note that each patch in MASH already provides a consistent normal orientation for the points inside, we only need to find a globally consistent orientation for all the patches. 
More specifically, for the $k$-th sampled point of $i$-th anchor $p^{i}_{k}$ in the sampled points $P$, we first obtain its normal $n_{\text{src}}$, 
with more details provided in the supplementary material.
Then we sample a subset of points $P'$ from $P$ by furthest point sampling and estimate their normals using PGR \cite{pgr}.
Once the globally consistent normals for the point set $P'$ are obtained, we reverse the normal direction of patches 
where the orientation is inconsistent with those points in $P'$, and then perform a smooth normal interpolation for better continuity between boundaries, defined as:
\begin{equation}
    n^{i}_{k} = \text{slerp}(n_{\text{src}}, n_{\text{pgr}},\sqrt{\omega^{i}_{k}} ),
\end{equation}
where $n_{\text{pgr}}$ is the normal of the nearest point $p' \in P'$ of $p^{i}_{k}$, and $\omega^{i}_{k} \in [0,1]$ is the corresponding mask parameter \cl{indicating} how $p^{i}_{k}$ is close to the center of the belonging patch.

\begin{table}[t]
\centering
\caption{
Quantitative comparison with surface reconstruction baselines 
on the ShapeNet-V2 dataset.
For ease of comparison of results, we multiply the L1 Chamfer Distance \rz{(L1-CD)} by 1000 here.
}
\label{tab:recon}
\resizebox{0.49\textwidth}{!}{
\iftrue
\begin{tabular}{ccccccc|c}
\toprule
Method&\cl{SPR}&PGR&\cl{CONet}&ARO&\cl{M+M}&\cl{MASH}&FPS\\
\hline
\cl{L1-CD}$\downarrow$&89.565&6.381&17.732&15.697&\underline{5.450}&\bf{4.944}&4.782\\
\cl{L2-CD}$\downarrow$&429.112&26.876&72.523&66.051&\underline{22.523}&\bf{2.268}&1.871\\
FScore$\uparrow$&0.497&0.988&0.812&0.880&\underline{0.997}&\bf{0.998}&0.999\\
$D_H$$\downarrow$&0.272&0.023&0.131&0.117&\underline{0.019}&\bf{0.013}&0.012\\
$S_{cos}$$\uparrow$&0.684&0.974&0.821&0.898&\underline{0.980}&\bf{0.984}&0.991\\
\cl{NIC}$\downarrow$&65.023&19.178&29.810&23.035&\underline{18.040}&\bf{13.346}&6.230\\
\bottomrule
\end{tabular}
\fi

}
\end{table}
\begin{table}[t]
\centering
\caption{
Quantitative comparison of category-conditioned generations.
}
\label{tab:category-gen}
\resizebox{0.36\textwidth}{!}{
\begin{tabular}{ccccc}
\toprule
        & R-KID $\downarrow$ & R-FID $\downarrow$ & P-KID $\downarrow$ & P-FID $\downarrow$ \\
\hline
LN3Diff & 0.222 & 247.379 & 0.792 & 150.578 \\
3DShape2VecSet & 0.138 & 173.649 & 0.146 & 38.021 \\
Ours & \bf{0.136} & \bf{163.473} & \bf{0.093} & \bf{30.344} \\
\bottomrule
\end{tabular}
}
\end{table}
\begin{table}[t]
\centering
\caption{
Quantitative comparison of image-conditioned generation.
}
\label{tab:image-gen}
\resizebox{0.47\textwidth}{!}{

\iftrue
\begin{tabular}{ccccccc}
\toprule
             & \cl{L1-CD} $\downarrow$ & R-KID $\downarrow$ & R-FID $\downarrow$ & P-KID $\downarrow$ & P-FID $\downarrow$ & ULIP-I $\uparrow$ \\
\hline
InstantMesh & 23.112 & 0.056 & 179.755 & 0.079 & 74.720 & 5.552 \\
Make-a-Shape & 31.105 & 0.032 & 161.428 & 0.074 & 49.857 & 7.360 \\
Hunyuan3D-1 & 11.227 & 0.023 & 160.322 & 0.036 & 51.534 & 6.367 \\
Ours & \bf{9.555} & \bf{0.018} & \bf{156.141} & \bf{0.005} & \bf{25.074} & \bf{7.530} \\
\bottomrule
\end{tabular}
\fi

}
\end{table}

\paragraph{Baseline comparisons}
\label{para:baseline}


We compare our method to Screened Poisson (\cl{SPR}) \cite{kazhdan2013screened} and several other representative reconstruction methods, including Parametric Gauss Reconstruction (PGR) \cite{pgr}, Convolutional Occupancy Networks (\cl{CONet}) \cite{conv-onet}, and ARO-Net \cite{aronet}.
PGR is the SOTA learning-free reconstruction method from point clouds without normals, while CONet and ARO are both learning-based. 
We conduct experiments on the full dataset of ShapeNet-V2~\cite{shapenet} \cl{with 8,192 input points as default}. 

\begin{figure*}[!t]
\includegraphics[width=0.85\textwidth]{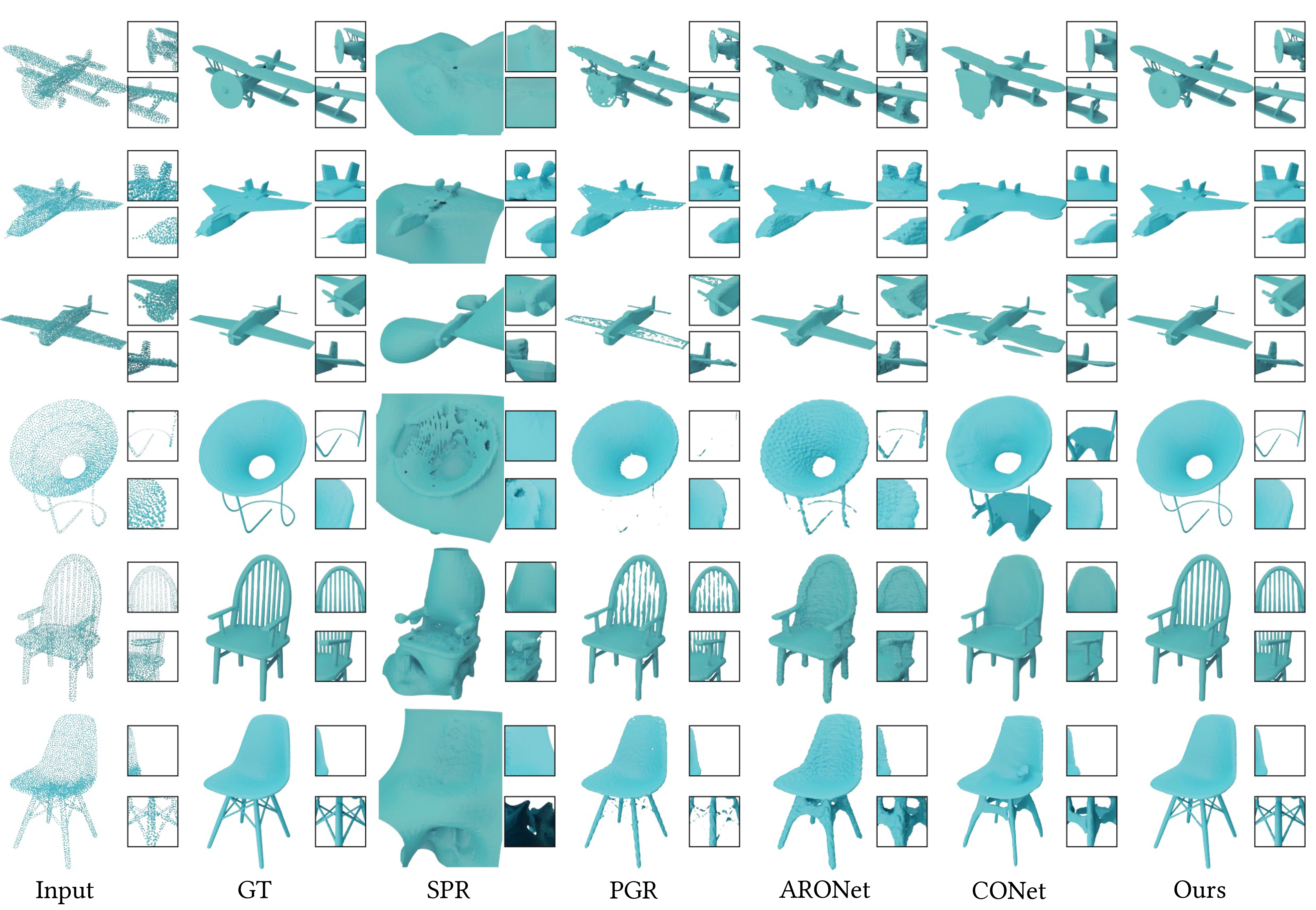}
\caption{\rz{Qualitative comparison on surface reconstruction.
Two zoom-ins are provided for each result to better show the geometric details.}
}
\label{fig:recon}
\end{figure*}

\begin{figure*}[!t]
\includegraphics[width=0.85\textwidth]{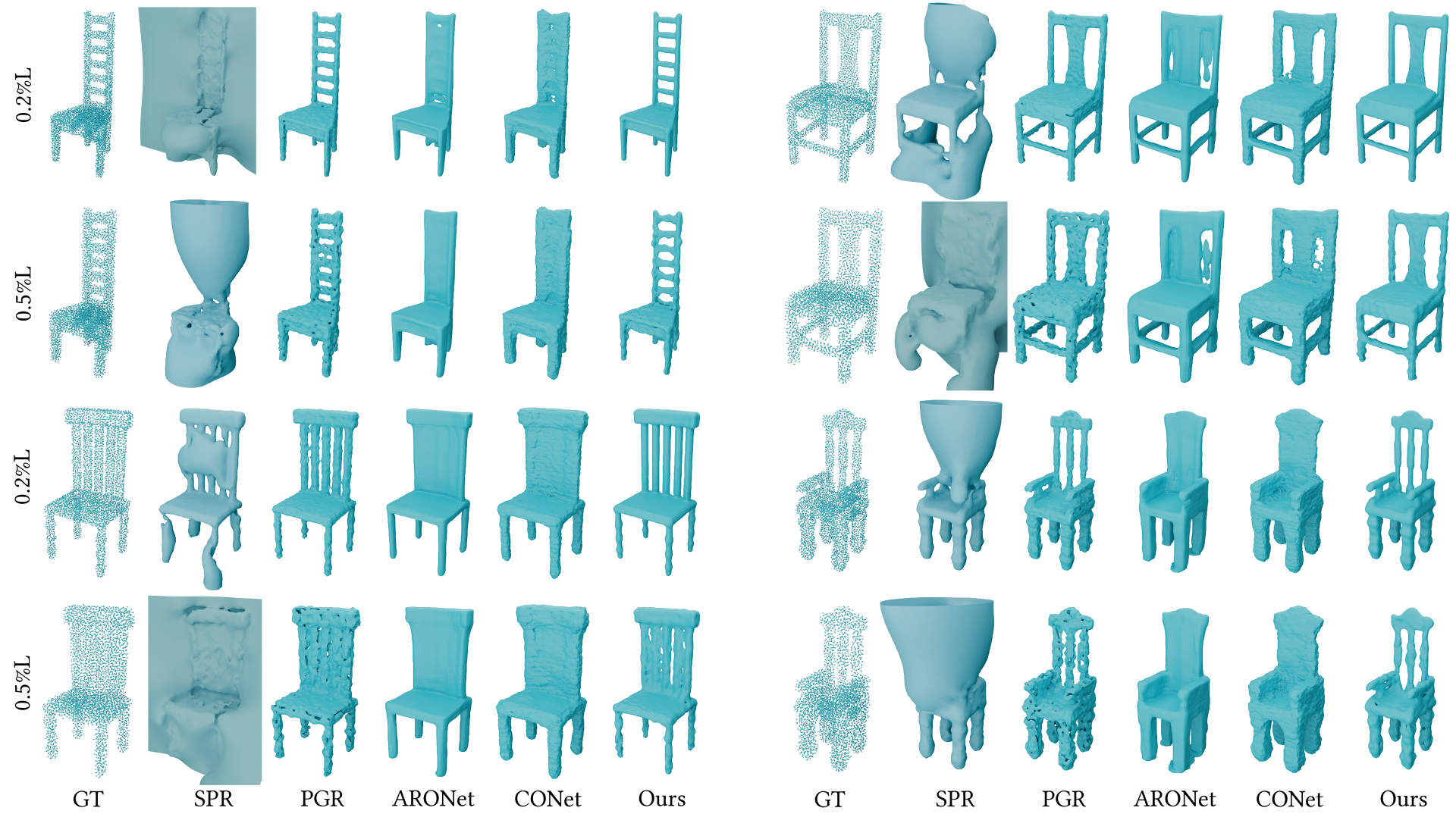}
\caption{Reconstructions on the chair category with two different noise levels.
}
\label{fig:noise}
\end{figure*}

Table~\ref{tab:recon} shows that our method outperforms all the baselines \rz{quantitatively}.
\rz{Qualitatively, as shown in Figure~\ref{fig:recon} with the zoom-ins, our MASH representation leads to clearly superior
surface reconstruction results in terms of smoothness quality, better conformation to the spatial distribution of the input point clouds, and in particular the ability to faithfully recover intricate geometric details (e.g., thin structures or dense grids).}
Our method is also more robust to noisy inputs, as demonstrated by representative results in Figure~\ref{fig:noise}.
\cl{
In addition, we have conducted further evaluations and comparisons on sparse, non-uniform data, as well as high-genus shapes.
}
More details about the experiment setup, evaluation metrics, results for per-category comparison, and explanations about the experiments on \cl{different settings} are provided in the supplementary material.

\subsection{Shape Generation}
\label{sec:gen}
\cl{
Since MASH is discriminative yet compact, it can naturally serve as embeddings for 3D shapes and has the potential to make learning and inference more efficient compared to implicit representations.
}
We show how MASH can be further used for shape generation in two most common settings, i.e., category-conditioned and image-conditioned. 
\cl{
Considering that MASH consists of a set of unordered surface patches, each represented by a fixed-length set of parameters, making it fit the architecture of 3DShape2VecSet~\cite{zhang20233dshape2vecset}. Therefore we also use the same conditioning for shape generation as in 3DShape2VecSet.
Specifically,} for the category-conditional model, we create a learnable embedding vector for each category and inject it as a condition into the network.
For the image-conditional model, we extract the image features using the pre-trained DINOv2-ViT-B/14 model \cite{oquab2023dinov2}. 
\cl{
To further improve the parameter distribution for generation, we applied an invertible piecewise linear transform to convert it into normal distributions, which is a modified version of the PowerTransformer in sklearn~\cite{sklearn_api} to ensure that the transformation error is below 1e-6 for better precision.
}
More details are provided in the Supplementary. 



\paragraph{Category-conditioned generation}
Our category-conditioned model is trained on ShapeNet-V2~\cite{shapenet}. We compare our method with LN3Diff~\cite{lan2025ln3diff} and 3DShape2VecSet (S2V) \cite{zhang20233dshape2vecset}.
Thanks to the compactness of MASH, the training of our network converges much quicker than other methods.
Compared to S2V, both the training and sampling time of our model is less than one-third, with the same network backbone. 

\begin{figure*}[!t]
\includegraphics[width=0.85\textwidth]{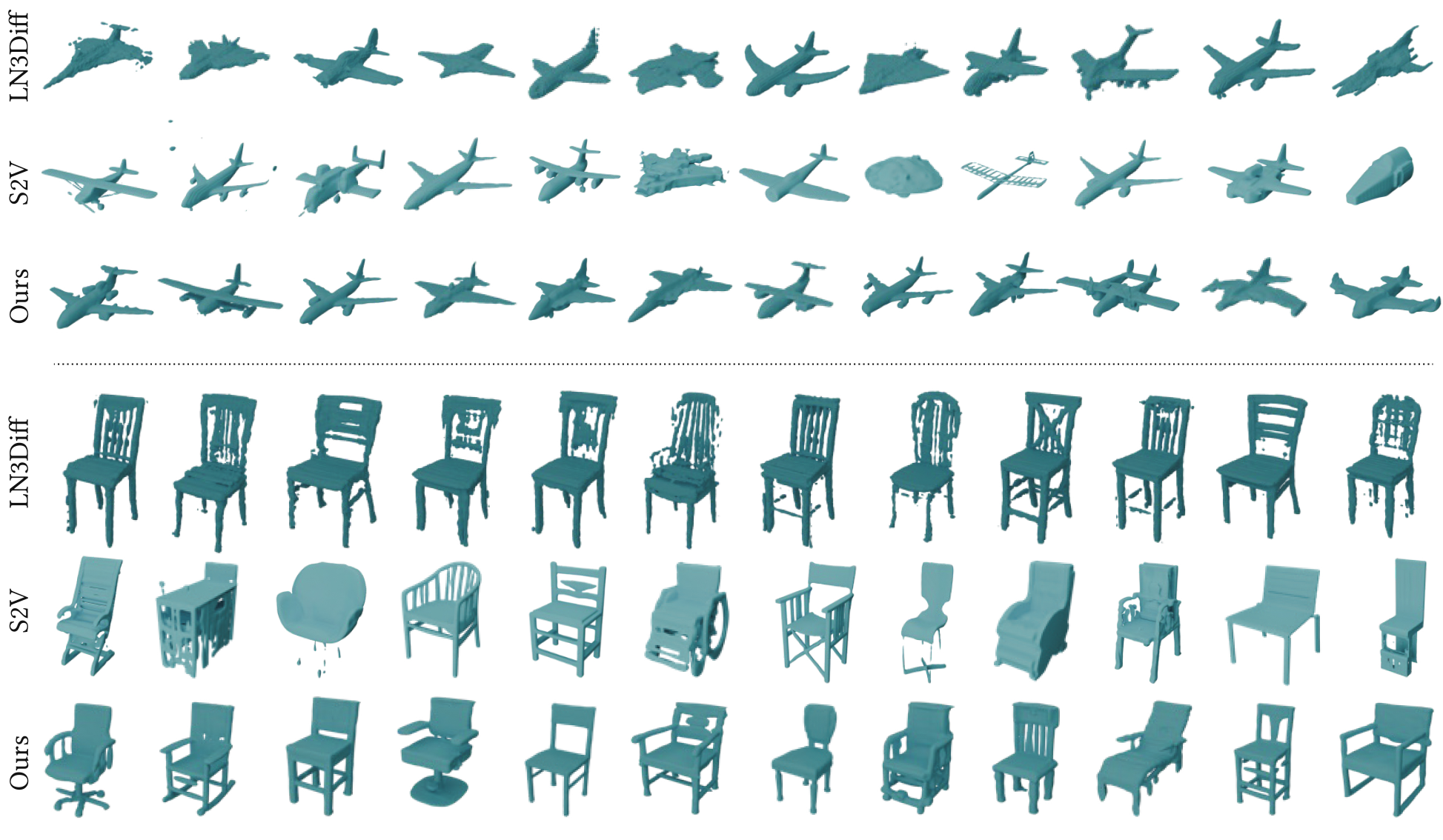}
\caption{Qualitative results on category-conditioned generation compared with different methods.
}
\label{fig:category-gen}
\end{figure*}

\rz{Following CLAY~\cite{zhang2024clay}, we measure quality of the generative meshes} using Render-FID, Render-KID, P-FID, and P-KID, computed on 200 generated shapes and shapes randomly selected from ShapeNet-V2 for each category.
Quantitative and qualitative comparisons are presented in Table~\ref{tab:category-gen} and Figure~\ref{fig:category-gen}, respectively.
\rz{The results clearly show that shapes generated via MASH exhibit greater diversity with better geometric details.}

\begin{figure}[!t]
\includegraphics[width=0.49\textwidth]{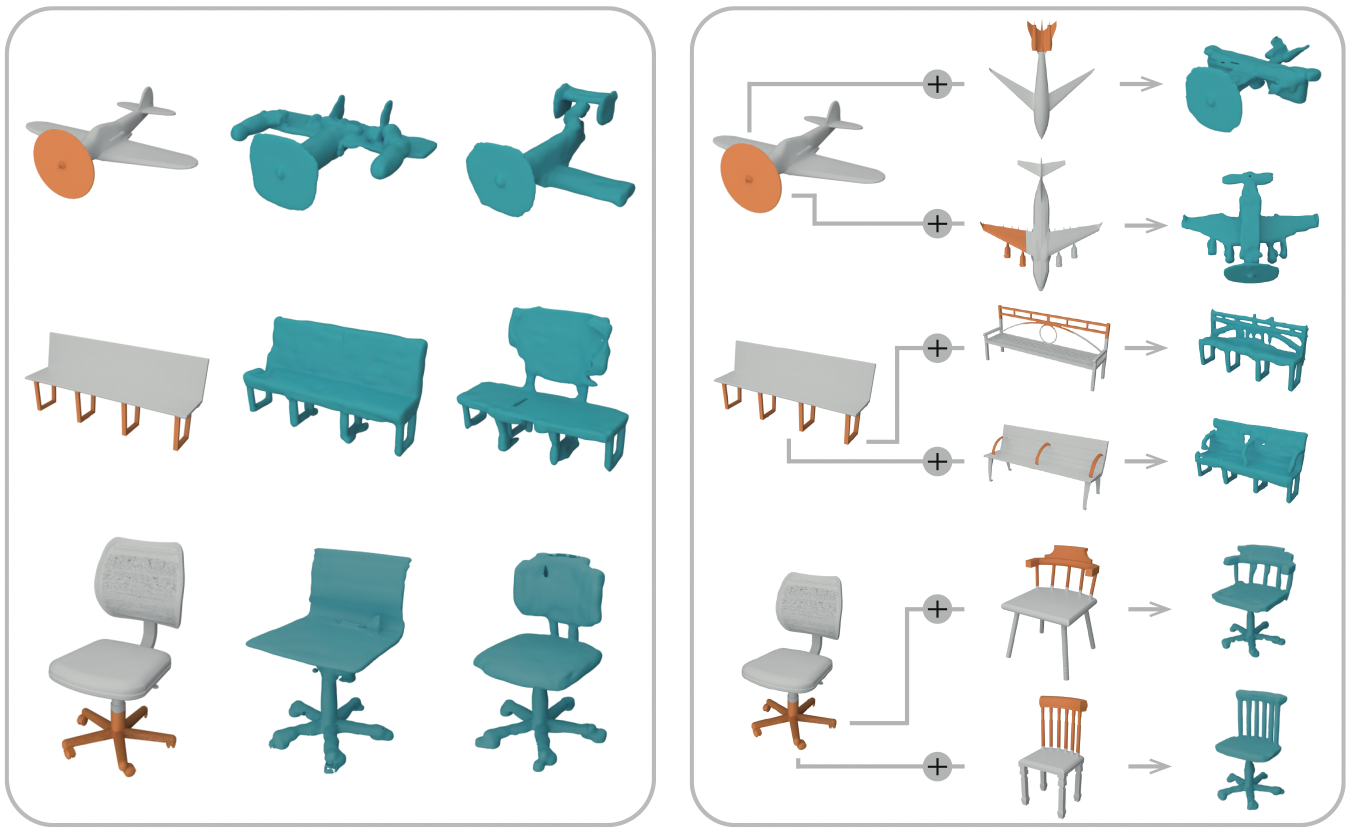}
\caption{Part-conditioned shape completion and blending enabled by our patch-based MASH representation. 
}
\label{fig:part-gen}
\end{figure}

Moreover, as an explicit patch-based representation, one unique feature of MASH is that a subset of its anchors can be fixed during the generation, which enables novel applications like part-conditional generation and shape blending. More specifically, 
during the training of our category-conditional model, we replace a random proportion of the initial noise fed into the generative model with the parameters of the ground truth MASH in 80\% of the steps. 
With this modification, our trained category-conditioned generative model can naturally support completion and blending tasks.
Some visual examples are shown in Figure~\ref{fig:part-gen}\rz{, where we use the orange color to highlight the retained shape parts, and cyan to represent the entire generated 3D shapes}. 
We can see that the generative model fully utilizes the information of the fixed anchors and attempts to provide reasonable completion or blending results, with
all fixed parts are nicely preserved. Note that the slight change of local geometry of the given part is mainly due to  surface extraction. 

\begin{figure*}[!t]
\includegraphics[width=0.98\textwidth]{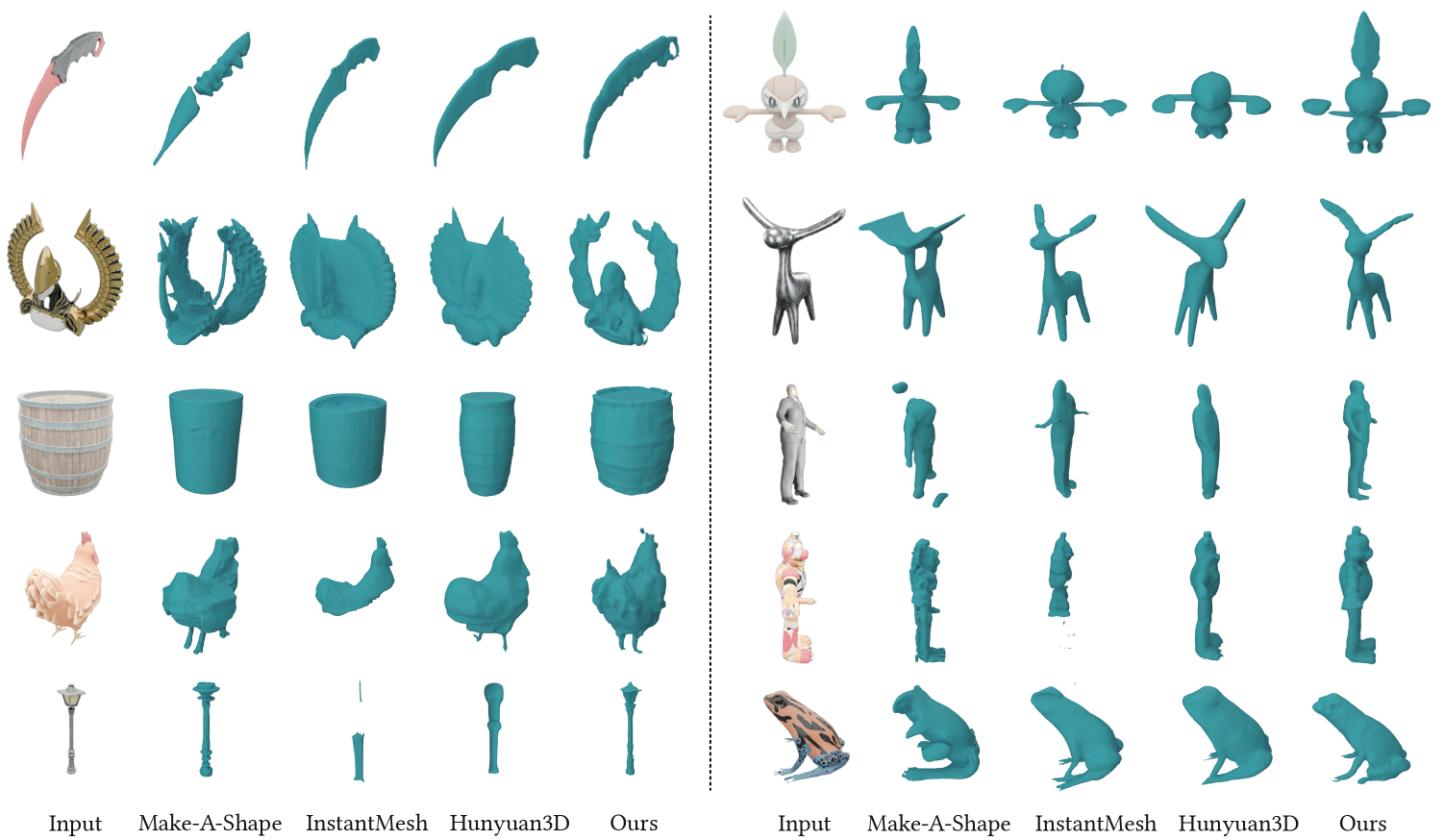}
\caption{Qualitative results on single image to 3D generation compared with different methods.
}
\label{fig:dino-gen}
\end{figure*}

\paragraph{Image-conditioned generation}
We train the image-conditioned model on the Objaverse-82K dataset~\cite{deitke2023objaverse} and compare our method to baselines including InstantMesh~\cite{xu2024instantmesh}, Make-A-Shape~\cite{hui2024make} and Hunyuan3D-1~\cite{yang2024hunyuan3d}.
To perform a quantitative comparison, we randomly select 100 3D shapes from the Objaverse dataset and render an image from a random viewpoint for each shape to serve as the input of different methods.
Then, we normalize the 3D shapes generated by different methods together with the 3D shapes in the dataset and manually align them to eliminate any possible orientation inconsistencies between the generated results and the ground truth. 
Other than the metrics used for category-conditioned generations, we also introduce Chamfer Distance (\cl{L1-CD}) and ULIP-I as additional metrics.
The quantitative and qualitative comparisons are shown in Table~\ref{tab:image-gen} and Figure~\ref{fig:dino-gen}, respectively.
Our method obtains generally better results.
Both InstantMesh and Hunyuan3D-1 take the input image to generate multi-view images first and then reconstruct the 3D shape from those images, thus they highly depend on the geometric consistency across multiple views.
Compared to other methods, the 3D shapes generated based on MASH can better preserve the topology of the object in the image and better recover the full geometric shape.

\section{Conclusion}

We introduce a new {\em multi-view\/} and {\em parametrized\/} 3D shape representation, coined Masked Anchored SpHerical Distances (MASH), to efficiently and accurately parameterize a given 3D shape. 
With a differentiable MASH optimization algorithm, we can convert discrete data structures such as point clouds into continuous surface-based representations efficiently. 
Moreover, we demonstrate that MASH is quite versatile for various applications including surface reconstruction, shape completion, blending, and generations.  
Extensive experiments and comparisons are conducted to validate the effectiveness of our new representation.

\paragraph{Limitation and future work}
As a preliminary attempt, our representation and corresponding optimization still have several limitations. 
Our current MASH representation only considers geometry and it would be interesting to explore ways to further incorporate textures.
For the optimization, the current initialization evenly distributes anchors over the given data, which sometimes lead to sub-optimal fitting results. 
Adapting the anchor distribution to the given shape will be a promising direction for future work.
\rz{
At last, since the current datasets employed for training our methods and the scale of the generative network are relatively small, the geometric details of the generated results still leave much room for improvements. Training a more powerful generative model on a larger-scale dataset is a worthwhile pursuit for future work.
}





\begin{acks}
\cl{We thank the anonymous reviewers for their valuable comments.
This work is supported by the
National Key R\&D Program of China (2024YFB2809102),
National Natural Science Foundation of China (U24B20154, 62025207, 62322207),
Joint NCSF-ISF Research Grant (3077/23),
and Shenzhen University Natural Sciences 2035 Program (2022C007).
}
\end{acks}

\bibliographystyle{ACM-Reference-Format}
\bibliography{reference}

\clearpage

\end{document}